\documentclass{article}

\usepackage{arxiv}

\usepackage{newtxtext}
\usepackage{newtxmath}

\usepackage[utf8]{inputenc}
\usepackage[T2A,T1]{fontenc}
\usepackage[russian,english]{babel}
\usepackage{hyperref}
\usepackage{url}
\usepackage{booktabs}
\usepackage{amsfonts}
\usepackage{nicefrac}
\usepackage{microtype}
\usepackage{cleveref}
\usepackage{lipsum}
\usepackage{graphicx}
\usepackage{natbib}
\usepackage{doi}

\newif\ifuniqueAffiliation
\uniqueAffiliationfalse

\title{Adapting Large Language Models to a Low-Resource Agglutinative Language:
A Comparative Study of LoRA and QLoRA for Bashkir}

\ifuniqueAffiliation
    \author{%
        \href{https://orcid.org/0000-0003-2525-1183}{\includegraphics[scale=0.06]{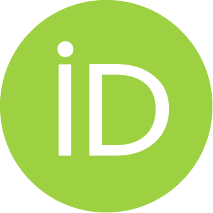}\hspace{1mm}M. K. Arabov}\thanks{Email: \texttt{MKArabov@kpfu.ru}}
        \And
        \href{https://orcid.org/0009-0003-8818-2370}{\includegraphics[scale=0.06]{orcid.pdf}\hspace{1mm}S. S. Khaybullina}\thanks{Email: \texttt{khaybulinas@mail.ru}} \\
        Institute of Computational Mathematics and Information Technologies\\
        Kazan Federal University\\
        Kazan, Russia \\
        \texttt{\{MKArabov, khaybulinas\}@kpfu.ru}
    }
\else
    \usepackage{authblk}
    
    \newbox{\orcid}\sbox{\orcid}{\includegraphics[scale=0.06]{orcid.pdf}}
    \author[1]{%
        \href{https://orcid.org/0000-0003-2525-1183}{\usebox{\orcid}\hspace{1mm}M. K. Arabov}\thanks{Email: \texttt{MKArabov@kpfu.ru}}%
    }
    \author[1]{%
        \href{https://orcid.org/0009-0003-8818-2370}{\usebox{\orcid}\hspace{1mm}S. S. Khaybullina}\thanks{Email: \texttt{khaybulinas@mail.ru}}%
    }
    \affil[1]{Institute of Computational Mathematics and Information Technologies, Kazan Federal University, Kazan, Russia}
\fi

\renewcommand{\shorttitle}{Adapting LLMs to Bashkir: LoRA and QLoRA Study}

\hypersetup{
    pdftitle={Adapting Large Language Models to a Low-Resource Agglutinative Language: A Comparative Study of LoRA and QLoRA for Bashkir},
    pdfsubject={cs.CL},
    pdfauthor={M. K. Arabov and S. S. Khaybullina},
    pdfkeywords={Bashkir language, LoRA, QLoRA, low-resource languages, parameter-efficient fine-tuning, large language models, comparative analysis},
}

\begin{document}
\maketitle

\begin{abstract}
This paper presents a comparative study of parameter-efficient fine-tuning (PEFT) methods, including LoRA and QLoRA, applied to the task of adapting large language models to the Bashkir language, a low-resource agglutinative language of the Turkic family. Experimental evaluation is conducted on a Bashkir text corpus of 71k documents (46.9M tokens) using models of various architectures: DistilGPT2, GPT-2 (base, medium), Phi-2, Qwen2.5-7B, DeepSeek-7B, and Mistral-7B. To improve the reliability of results, each configuration was trained with three different random seeds.

The lowest perplexity on the test set was obtained for GPT-2 medium with full fine-tuning (3.34). Meanwhile, QLoRA applied to Mistral-7B (3.79) and Phi-2 (3.81) achieved comparable quality with over 40 times fewer trainable parameters. However, we also observed cases of significant quality degradation when using PEFT for certain architectures (e.g., DeepSeek-7B with rank 8, perplexity = 129.55), indicating that the outcome depends critically on the choice of the base model and its tokenizer.

Additionally, a qualitative analysis of generated texts based on Bashkir prompts revealed that models with the best perplexity do not necessarily produce the most coherent outputs: QLoRA-tuned models generated monolingual Bashkir continuations, whereas the fully fine-tuned model with the lowest perplexity frequently switched to English. The results suggest that QLoRA on 7B-scale models offers an effective compromise between quality and computational cost for Bashkir. To ensure reproducibility, open data, code, and trained adapters will be released upon acceptance.
\end{abstract}


\section{Introduction}

The rapid advancement of large language models (LLMs), built upon the foundational Transformer architecture~\cite{vaswani2017attention}, has established them as fundamental tools across a wide spectrum of natural language processing (NLP) applications. However, the performance of these models is critically dependent on the volume and quality of training data, a circumstance that severely restricts their applicability to languages with limited digital representation. The Bashkir language, a member of the Turkic language family characterized by agglutinative morphology and a Cyrillic-based script, exemplifies such a low-resource setting. Despite having several million speakers, the scarcity of publicly available textual resources for Bashkir impedes the effective application of modern NLP techniques.

This challenge is compounded by the fact that most state-of-the-art LLMs are predominantly trained on high-resource languages, particularly English. Consequently, their direct application to Bashkir text yields suboptimal performance. While full fine-tuning of even moderately sized models can mitigate this issue, it entails substantial computational costs that limit its practical feasibility. In this context, parameter-efficient fine-tuning (PEFT) methods have emerged as a compelling alternative, aiming to reduce computational burdens while preserving competitive performance levels. Among these, Low-Rank Adaptation (LoRA)~\cite{hu2022lora} is a widely adopted technique that freezes the original model parameters and introduces trainable low-rank matrices into specific layers. A subsequent advancement, QLoRA~\cite{dettmers2023qlora}, further enhances efficiency by integrating low-rank adaptation with low-bitweight quantization, drastically reducing memory and computational requirements.

Despite the active development of PEFT methods and their potential for low-resource language scenarios, as documented in recent comprehensive surveys~\cite{han2024peftsurvey,xu2024peftsurvey}, comparative studies analyzing the effectiveness of LoRA and QLoRA for the Bashkir language across diverse LLM architectures remain scarce. The present work aims to fill this gap by providing a systematic experimental evaluation of these methods on a corpus of Bashkir texts.

\section{Related Work}

The present study sits at the intersection of three research streams: parameter-efficient fine-tuning methods, their emerging application to low-resource and Turkic languages, and foundational work on tokenization strategies for morphologically rich languages.

\subsection*{Parameter-Efficient Fine-Tuning: From LoRA to QLoRA}

The prohibitive cost of fully fine-tuning large language models has driven the development of parameter-efficient alternatives. LoRA (Low-Rank Adaptation), introduced by Hu et al.~\cite{hu2022lora}, decomposes weight updates into products of low-rank matrices, reducing the number of trainable parameters by orders of magnitude while preserving competitive downstream performance. QLoRA~\cite{dettmers2023qlora} extended this paradigm by coupling low-rank adaptation with 4-bit NormalFloat quantization, demonstrating that models with up to 65B parameters can be fine-tuned on a single GPU without sacrificing quality. As PEFT methods have proliferated, comprehensive surveys by Han et al.~\cite{han2024peftsurvey} and Xu et al.~\cite{xu2024peftsurvey} have systematically categorized available approaches---including adapters, prefix tuning, and prompt tuning alongside LoRA variants---and distilled practical guidance for method selection. However, these surveys also reveal a striking imbalance: the vast majority of empirical PEFT studies evaluate on English and a handful of other high-resource languages, leaving the behavior of these methods on linguistically diverse, low-resource languages largely uncharted.

\subsection*{PEFT for Low-Resource and Turkic Languages}

A nascent but growing body of work has begun applying PEFT to languages with limited digital representation. Khade et al.~\cite{khade2024challenges} showed that LoRA-based fine-tuning of multilingual LLMs substantially outperforms full fine-tuning for Marathi, an Indo-Aryan language with moderate resources, when training data is scarce. Joshi et al.~\cite{joshi2024finetuning} provided a broader comparative study across multiple low-resource Indian languages, confirming the advantage of PEFT in low-data regimes and noting that the optimal configuration varies considerably across target languages.

Within the Turkic family, the challenge of resource scarcity is particularly acute for languages other than Turkish. Baghirova et al.~\cite{baghirova2024kardes} developed the Karde\c{s}-NLU benchmark, a suite of natural language understanding tasks for five Turkic languages, and demonstrated that transfer learning from Turkish yields substantial improvements on lower-resource relatives. In automatic speech recognition, Mamyrbayev et al.~\cite{mamyrbayev2024multilingual} trained multilingual end-to-end models on five Turkic languages with shared Cyrillic orthographies and found that Bashkir exhibited consistently higher word error rates than Kazakh, Tatar, or Kyrgyz, suggesting that certain phonological or lexical properties of Bashkir pose unique modeling challenges.

Most directly relevant to our work, Karpov~\cite{karpov2026translation} applied LoRA to fine-tune the NLLB-200 model for Russian--Bashkir machine translation, achieving a chrF++ score of 46.94. This study constitutes the first and, to our knowledge, only prior application of PEFT specifically to Bashkir. However, it is confined to a single model (NLLB-200), a single task (machine translation), and does not systematically compare alternative methods, ranks, or architectures. A comprehensive investigation of how different PEFT strategies affect generative language modeling quality for Bashkir across diverse LLM architectures remains entirely absent from the literature.

\subsection*{Tokenization and the Challenge of Agglutinative Morphology}

A growing consensus holds that tokenizer quality is a decisive factor in cross-lingual transfer, particularly for languages that are underrepresented in pre-training corpora. Rust et al.~\cite{rust2021tokenizer} provided a systematic evaluation showing that multilingual tokenizers often fragment words in lower-resource languages into excessively long subword sequences, degrading downstream task performance. This problem is particularly severe for agglutinative languages, where a single word can encode what would be a multi-word phrase in English, as demonstrated by Sennrich et al.~\cite{sennrich2016bpe} in their foundational work introducing Byte Pair Encoding (BPE) for neural machine translation. In such languages, each token may carry multiple grammatical morphemes, and suboptimal segmentation directly impacts the model's ability to learn coherent representations.

These tokenization challenges have direct and underexplored implications for LLM adaptation to Bashkir. As an agglutinative Turkic language written in Cyrillic script, Bashkir combines two properties known to challenge multilingual tokenizers: complex morphological structure and a script underrepresented in pre-training data. Prior work on Turkic languages~\cite{mamyrbayev2024multilingual} has already identified Bashkir as more difficult to model than its closer relatives, and tokenization quality is a plausible contributing factor. However, no study to date has systematically investigated how tokenizer behavior interacts with PEFT configuration for Bashkir generative language modeling.

\subsection*{Research Gap and Present Contribution}

In summary, the literature reveals a clear and consequential gap. PEFT methods are well-studied theoretically~\cite{han2024peftsurvey,xu2024peftsurvey} and have shown promise for several low-resource languages~\cite{khade2024challenges,joshi2024finetuning,baghirova2024kardes}. Within the Turkic family, Bashkir has been identified as particularly challenging for multilingual models~\cite{mamyrbayev2024multilingual}, yet prior work applying PEFT to Bashkir is limited to a single study on machine translation with one model~\cite{karpov2026translation}. No research to date has systematically compared LoRA and QLoRA across multiple model architectures and ranks for generative language modeling in Bashkir, nor has any study systematically analyzed how tokenizer behavior may interact with PEFT effectiveness for an agglutinative Turkic language. The present work addresses this gap through a rigorous experimental evaluation spanning 7 models, 3 fine-tuning strategies, 2 ranks, and multiple random seeds, complemented by both quantitative and qualitative analysis.

\section{Experimental Setup}

\subsection{Data}

For this study, a Bashkir text corpus was constructed by aggregating datasets from 16 open sources. The resulting corpus contains 71,567 documents totaling approximately 46.9 million tokens.\footnote{The corpus will be made publicly available under open licenses upon acceptance.}

The corpus encompasses texts of diverse genres and functional styles, including encyclopedic articles, news materials, literary journal publications, user-generated content, and digitized book editions. The distribution of documents across sources is heterogeneous, with encyclopedic texts constituting the largest share, while other data types are represented in smaller proportions, reflecting the current state of digital resources for the Bashkir language.

During preprocessing, textual content was extracted from each document, followed by length unification through truncation to a maximum of 2,000 characters. Due to computational constraints, a random subsample of 10,000 documents was drawn for the experiments. This dataset was split into training, validation, and test sets in an 80:10:10 ratio. The split was performed while preserving the proportional distribution of sources. To ensure reproducibility, a fixed random seed was used for data partitioning.

\subsection{Models and Fine-Tuning Methods}

For the comparative analysis, seven base architectures of large language models were selected, covering a wide range of sizes and pre-training paradigms. The smaller-scale group comprised three decoder-only Transformer models: the distilled 82M-parameter DistilGPT2, the 124M GPT-2 base, and the 355M GPT-2 medium. The large-scale group consisted of four models in the multi-billion parameter range: Phi-2 (2.7B), a compact model optimized for reasoning tasks; Qwen2.5-7B-Instruct (7B), a general-purpose instruction-tuned model; DeepSeek-LLM-7B-Chat (7B), a dialogue-oriented model; and Mistral-7B-v0.3 (7B), featuring an improved modern architecture. A summary of all models is given in Table~\ref{tab:models}.

\begin{table}[ht]
\centering
\caption{Models used in the experiments}
\label{tab:models}
\begin{tabular}{lr}
\hline
\textbf{Model} & \textbf{Parameters} \\
\hline
DistilGPT2            & 82M \\
GPT-2                 & 124M \\
GPT-2 medium          & 355M \\
Phi-2                 & 2.7B \\
Qwen2.5-7B-Instruct   & 7B \\
DeepSeek-LLM-7B-Chat  & 7B \\
Mistral-7B-v0.3       & 7B \\
\hline
\end{tabular}
\end{table}

Three fine-tuning strategies were evaluated:

\begin{itemize}
    \item \textbf{Full fine-tuning (baseline)} --- updating all model parameters on the training data. This strategy was applied to all models except Phi-2 (2.7B), whose full fine-tuning was not conducted due to computational constraints.
    
    \item \textbf{LoRA} --- parameter-efficient fine-tuning using low-rank adapters~\cite{hu2022lora}. The experiments employed ranks $r=8$ and $r=16$, with the scaling factor set as $\alpha=2r$ and a dropout value of 0.05. For GPT-2 family models, adaptation was applied to the attention module, whereas for other architectures it targeted the self-attention projections. For Mistral, modules associated with hidden state transformations were additionally adapted.
    
    \item \textbf{QLoRA} --- an extension of LoRA combining low-rank adaptation with 4-bit weight quantization~\cite{dettmers2023qlora}. This approach was applied to models with 2.7B parameters and above: Phi-2, Qwen2.5-7B, DeepSeek-7B, and Mistral-7B.
\end{itemize}

All experiments were conducted on the 10,000-document subsample. For each combination of model, fine-tuning method, and rank, training was performed with three different random seeds, with the exception of DeepSeek-7B with rank~8, where the third run was terminated due to anomalously high perplexity (see Subsection~\ref{subsec:quantitative}).

\subsection{Training Hyperparameters}

Model training was carried out using the Hugging Face Transformers and PEFT libraries. The maximum sequence length was set to 128 tokens. The optimizer was AdamW with a learning rate of $2 \times 10^{-4}$, and cross-entropy was used as the loss function. Gradient accumulation was performed over 2 steps. The number of training epochs was set to 3 for models up to 2.7B parameters and 2 for 7B models.

Batch sizes varied according to model scale due to GPU memory constraints, as summarized in Table~\ref{tab:hyperparams}.

\begin{table}[ht]
\centering
\caption{Training hyperparameters}
\label{tab:hyperparams}
\begin{tabular}{ll}
\hline
\textbf{Parameter} & \textbf{Value} \\
\hline
Sequence length      & 128 tokens \\
Learning rate        & $2 \times 10^{-4}$ \\
Gradient accum.      & 2 steps \\
Epochs (up to 2.7B) & 3 \\
Epochs (7B)          & 2 \\
Batch size (DistilGPT2, GPT-2) & 8 \\
Batch size (GPT-2 medium)      & 4 \\
Batch size (2.7B--7B)          & 1 \\
Optimizer            & AdamW \\
Loss function        & Cross-entropy \\
\hline
\end{tabular}
\end{table}

The primary evaluation metric was perplexity, computed on the test set. Additionally, training time was recorded for each configuration.

\subsection{Qualitative Evaluation}

Beyond quantitative assessment, an analysis of generated text quality was conducted. For this purpose, the four models achieving the best perplexity scores were selected: GPT-2 medium with full fine-tuning, Mistral-7B with QLoRA ($r{=}16$), Phi-2 with QLoRA ($r{=}8$), and GPT-2 base with full fine-tuning. Each model was provided with a set of 10 short textual prompts in the Bashkir language.

Generation was performed in a sequence continuation mode without the use of additional control instructions. Greedy decoding was employed as the decoding method. The generated texts were analyzed according to the following criteria: grammatical correctness, semantic coherence, topical relevance to the prompt, and absence of language switching into other languages. The qualitative analysis complemented the quantitative results and revealed behavioral characteristics of the models that are not captured by automatic evaluation metrics.

\section{Results}

This section presents the results of the experimental comparison of various fine-tuning strategies for large language models on the Bashkir language. The evaluation was conducted along three main dimensions: model quality, measured by perplexity on the test set (Subsection~\ref{subsec:quantitative}); computational efficiency, characterized by the number of trainable parameters and training time (Subsection~\ref{subsec:efficiency}); and a qualitative analysis of generated texts on typical Bashkir prompts (Subsection~\ref{subsec:qualitative}). Additionally, Subsection~\ref{subsec:rank} analyzes the impact of the low-rank adaptation rank on final perplexity.

\subsection{Quantitative Comparison of Model Quality}
\label{subsec:quantitative}

The first stage of the analysis consisted of computing the mean perplexity for each configuration on the test set of 1,000 documents (10\% of the total subsample). To improve result reliability, each configuration was trained with three different random seeds (42, 43, 44), after which the arithmetic mean and standard deviation were calculated. The results are presented in Tables~\ref{tab:results_best} and~\ref{tab:results_all}, where models are sorted by increasing mean perplexity (lower values indicate better language model quality).

\begin{table}[ht]
\centering
\caption{Top-performing configurations by perplexity on the test set (mean $\pm$ std)}
\label{tab:results_best}
\begin{tabular}{llr}
\hline
\textbf{Model} & \textbf{Method} & \textbf{Perplexity} \\
\hline
GPT-2 medium & Full FT & $3.34 \pm 0.01$ \\
Mistral-7B   & QLoRA ($r=16$) & $3.79 \pm 0.02$ \\
Phi-2        & QLoRA ($r=8$)  & $3.81 \pm 0.03$ \\
Mistral-7B   & QLoRA ($r=8$)  & $3.85 \pm 0.01$ \\
GPT-2 base   & Full FT & $4.00 \pm 0.04$ \\
DistilGPT2   & Full FT & $4.42 \pm 0.02$ \\
\hline
\end{tabular}
\end{table}

\begin{table}[ht]
\centering
\caption{Remaining configurations by perplexity (mean $\pm$ std)}
\label{tab:results_all}
\begin{tabular}{llr}
\hline
\textbf{Model} & \textbf{Method} & \textbf{Perplexity} \\
\hline
Qwen2.5-7B & QLoRA, $r{=}16$ & $5.07 \pm 0.03$ \\
Qwen2.5-7B & QLoRA, $r{=}8$  & $5.34 \pm 0.04$ \\
GPT-2 med. & LoRA, $r{=}16$  & $5.93 \pm 0.01$ \\
GPT-2 med. & LoRA, $r{=}8$   & $6.53 \pm 0.00$ \\
GPT-2 base & LoRA, $r{=}16$  & $8.75 \pm 0.03$ \\
DeepSeek-7B& QLoRA, $r{=}16$ & $8.98 \pm 0.10$ \\
GPT-2 base & LoRA, $r{=}8$   & $9.49 \pm 0.03$ \\
DistilGPT2 & LoRA, $r{=}16$  & $10.27 \pm 0.03$ \\
DistilGPT2 & LoRA, $r{=}8$   & $11.13 \pm 0.02$ \\
DeepSeek-7B& QLoRA, $r{=}8$  & $129.55 \pm 1.06$ \\
\hline
\end{tabular}
\end{table}

As can be seen from Table~\ref{tab:results_best}, the best result is achieved with full fine-tuning of GPT-2 medium (perplexity 3.34). The standard deviation is only 0.01, indicating high training stability across different random initializations. The next best configurations are QLoRA applied to Mistral-7B ($r=16$, perplexity 3.79) and Phi-2 ($r=8$, perplexity 3.81). The difference between the leader and these two configurations is approximately 12--14\%, which, at such low absolute perplexity values, can be considered negligible. It is particularly noteworthy that the QLoRA versions of Mistral and Phi-2 achieve nearly the same quality as full fine-tuning of the more than two times smaller GPT-2 medium, while training only around 1--8 million parameters instead of 355 million.

Full fine-tuning of the base GPT-2 family models (GPT-2 base and DistilGPT2) yields perplexities of 4.00 and 4.42, respectively, which is expectedly worse than the larger GPT-2 medium but still remains at a perfectly acceptable level for practical applications. In contrast, the application of standard LoRA (without quantization) to the same models leads to substantially higher perplexity: for GPT-2 medium, LoRA with $r=8$ yields 6.53, and with $r=16$ yields 5.93, which is 1.7--2 times higher than full fine-tuning. A similar pattern is observed for GPT-2 base and DistilGPT2. This suggests that for relatively small models (under 1B parameters) and limited training data (10k documents), low-rank approximation without quantization proves insufficiently expressive to fully capture the lexical and grammatical specificities of the Bashkir language.

The anomalously high perplexity values for DeepSeek-7B with rank 8 (129.55) warrant separate discussion.\footnote{Due to the anomalously high perplexity rendering further training meaningless, the third run with this rank was terminated; therefore, only results from two random initializations are reported for this configuration.} Manual inspection of the generated texts revealed that the model virtually failed to learn coherent Bashkir phrases, frequently outputting sequences of special symbols or switching to English. The likely cause is the incompatibility of the original DeepSeek tokenizer with Bashkir Cyrillic: many Bashkir characters may have been absent from the vocabulary or represented by rare tokens, which, at a low adapter rank, prevented the embeddings from being sufficiently corrected. Increasing the rank to 16 significantly improved the situation (perplexity dropped to 8.98); however, the quality still lags behind other 7B models, such as Mistral (3.79) and even Qwen2.5 (5.07). This result underscores the critical importance of base model selection and its tokenizer when working with low-resource languages distinct from English.

\subsection{Computational Efficiency}
\label{subsec:efficiency}

Beyond model quality, resource consumption is a crucial criterion when selecting a fine-tuning strategy. Table~\ref{tab:efficiency} presents, for selected configurations, two indicators: the total number of trainable parameters (including adapter parameters for LoRA/QLoRA) and the training time over three epochs.

\begin{table}[ht]
\centering
\caption{Trainable parameters and training time}
\label{tab:efficiency}
\begin{tabular}{lcrr}
\hline
\textbf{Model} & \textbf{Method} & \textbf{Trainable (M)} & \textbf{Time (s)} \\
\hline
DistilGPT2   & Full FT         & 82.0   & 24.3 \\
DistilGPT2   & LoRA, $r{=}8$   & 0.3    & 17.8 \\
GPT-2 base   & Full FT         & 124.0  & 39.4 \\
GPT-2 base   & LoRA($r{=}8$)   & 0.4    & 31.8 \\
GPT-2 med.   & Full FT         & 355.0  & 130.1 \\
GPT-2 med.   & LoRA($r{=}8$)   & 0.9    & 126.2 \\
Phi-2        & QLoRA($r{=}8$)  & 1.7    & 429.7 \\
Mistral-7B   & QLoRA($r{=}16$) & 8.4    & 2001.8 \\
\hline
\end{tabular}
\end{table}

Full fine-tuning of all layers requires updating hundreds of millions of parameters even for relatively small models (82--355M). The application of LoRA/QLoRA reduces this number by orders of magnitude: for DistilGPT2 with $r{=}8$, only about 0.3M parameters are trained; for GPT-2 medium, 0.9M; for Phi-2, 1.7M; and for Mistral-7B, 8.4M. At the same time, as shown in Table~\ref{tab:results_best}, the quality of QLoRA-tuned Phi-2 and Mistral is virtually on par with full fine-tuning of GPT-2 medium, and in terms of generation stability even surpasses it (see Subsection~\ref{subsec:qualitative}).

Notably, for smaller models (DistilGPT2, GPT-2 base, GPT-2 medium), the reduction in training time when using LoRA is less pronounced: the difference ranges from 4\% (for GPT-2 medium) to 27\% (for DistilGPT2). This is because the dominant contribution to training time comes from the forward pass through all model layers, which is performed regardless of whether all weights or only adapters are updated. For large models (Phi-2, Mistral-7B), the absolute training time is considerably higher; however, the use of parameter-efficient fine-tuning methods such as QLoRA enables training models with billions of parameters on accessible computational resources. Thus, QLoRA provides a practical compromise between quality and computational cost, particularly for large-scale models.

\subsection{Qualitative Analysis of Generated Texts}
\label{subsec:qualitative}

Quantitative metrics such as perplexity do not always fully reflect a model's suitability for real-world text generation tasks. A qualitative analysis was therefore conducted: the four models with the best perplexity scores---GPT-2 medium with full fine-tuning, Mistral-7B with QLoRA ($r{=}16$), Phi-2 with QLoRA ($r{=}8$), and GPT-2 base with full fine-tuning---were presented with ten short incomplete phrases in Bashkir. Generation was performed in continuation mode without additional instructions, using greedy decoding. Representative examples are shown in Table~\ref{tab:generation}.

\begin{table*}
\centering
\caption{Generation examples on Bashkir prompts. QLoRA models use the best-performing ranks: $r{=}16$ for Mistral-7B and $r{=}8$ for Phi-2.}
\label{tab:generation}
\small
\raggedright
\begin{tabular}{| p{2.5cm} | p{3.0cm} | p{3.0cm} | p{3.0cm} | p{3.0cm} |}
\hline
\textbf{Prompt} & \textbf{GPT-2 med. (FT)} & \textbf{Mistral (QLoRA)} & \textbf{Phi-2 (QLoRA)} & \textbf{GPT-2 base (FT)} \\
\hline

{\fontencoding{T2A}\selectfont Һаумы, һин кем?} &

{\fontencoding{T2A}\selectfont Һаумы, һин кем?} \newline 
{\fontencoding{T2A}\selectfont Отворовлюдь: (кся) :} \newline 
I'm sorry... &

{\fontencoding{T2A}\selectfont Һаумы, һин кем?} \newline 
{\fontencoding{T2A}\selectfont Уның хәрби менән} \newline 
{\fontencoding{T2A}\selectfont уйлап сығыуым,} \newline 
{\fontencoding{T2A}\selectfont милли хәрбиәттең} \newline 
{\fontencoding{T2A}\selectfont барл...} &

{\fontencoding{T2A}\selectfont Һаумы, һин кем?} \newline 
{\fontencoding{T2A}\selectfont Һаумында айырып,} \newline 
{\fontencoding{T2A}\selectfont һинең үҙеге бар...} &

{\fontencoding{T2A}\selectfont Һаумы, һин кем?} \newline 
{\fontencoding{T2A}\selectfont Оівлядьстор. (1)} \newline 
We're going to get... \\
\hline

{\fontencoding{T2A}\selectfont Өфө —} &

{\fontencoding{T2A}\selectfont Өфө —} \newline 
{\fontencoding{T2A}\selectfont написок редоны,} \newline 
{\fontencoding{T2A}\selectfont 2ут 5-16.} I think \newline 
it is time... &

{\fontencoding{T2A}\selectfont Өфө — Мәскәүҙән} \newline 
{\fontencoding{T2A}\selectfont ауыр яуа} \newline 
{\fontencoding{T2A}\selectfont Өфөлә Мәскәүҙән} \newline 
{\fontencoding{T2A}\selectfont ауыр яуа —} \newline 
{\fontencoding{T2A}\selectfont Төркияла,} \newline 
{\fontencoding{T2A}\selectfont Ысәмбул...} &

{\fontencoding{T2A}\selectfont Өфө — Бөрйән} \newline 
{\fontencoding{T2A}\selectfont метро системы} \newline 
{\fontencoding{T2A}\selectfont ҡушылды} \newline 
{\fontencoding{T2A}\selectfont Өфө метро\-} \newline 
{\fontencoding{T2A}\selectfont система\-һының...} &

{\fontencoding{T2A}\selectfont Өфө —} \newline 
mòngs ómån (1) \newline 
A person who is \newline 
an inhabitant... \\
\hline

{\fontencoding{T2A}\selectfont Башҡортостан} &

{\fontencoding{T2A}\selectfont Башҡортостания} \newline 
{\fontencoding{T2A}\selectfont одоблемы уква,} \newline 
``a few thousand \newline 
Russians hav...'' &

{\fontencoding{T2A}\selectfont Башҡортостанда} \newline 
{\fontencoding{T2A}\selectfont уйынсыз} \newline 
{\fontencoding{T2A}\selectfont күңеленә} \newline 
{\fontencoding{T2A}\selectfont Башҡортостанның} \newline 
{\fontencoding{T2A}\selectfont Тәтешле районы} \newline 
{\fontencoding{T2A}\selectfont мәхкәм...} &

{\fontencoding{T2A}\selectfont Башҡортостанда} \newline 
{\fontencoding{T2A}\selectfont 50 мең кеше} \newline 
{\fontencoding{T2A}\selectfont ярҙамын итә} \newline 
{\fontencoding{T2A}\selectfont Бөгөн, 24 июлдә,} \newline 
{\fontencoding{T2A}\selectfont Башҡортостанд...} &

{\fontencoding{T2A}\selectfont Башҡортостания} \newline 
{\fontencoding{T2A}\selectfont коглеку, ьвы.} \newline 
Makka Kudy \newline 
(born on 12 \newline 
January 1927)... \\
\hline
\end{tabular}
\end{table*}

As can be seen from Table~\ref{tab:generation}, the fully fine-tuned GPT-2 medium often starts generation with meaningful Bashkir words (e.g., {\fontencoding{T2A}\selectfont ``Һаумы, һин кем?''}), but almost always then switches to English or produces incoherent symbol sequences. GPT-2 base exhibits an even stronger tendency toward language switching already at the first or second token after the prompt. In contrast, the models fine-tuned with QLoRA---Mistral-7B and Phi-2---generate fully coherent continuations in Bashkir, preserving grammatical structure (agglutinative forms, case endings) and topical relevance. For example, on the prompt {\fontencoding{T2A}\selectfont ``Өфө —''} (Ufa~---), Mistral continues with {\fontencoding{T2A}\selectfont ``Мәскәүҙән ауыр яуа''} (a heavy battle from Moscow), which is semantically connected to political discourse, while Phi-2 generates {\fontencoding{T2A}\selectfont ``Бөрйән метро системы ҡушылды''} (connected to the Burzyan metro system)---a grammatically correct, albeit factually unsupported, phrase.

It is important to note that the Mistral model with QLoRA never switched to another language across all tested prompts, and Phi-2 exhibited only isolated insertions of Russian words. Thus, the qualitative analysis confirms that for the low-resource Bashkir language, QLoRA on modern 7B-scale models ensures not only low perplexity but also high semantic coherence and linguistic purity.

\subsection{Impact of LoRA/QLoRA Rank}
\label{subsec:rank}

Finally, the impact of the low-rank adaptation rank on final quality was investigated. Table~\ref{tab:rank} presents perplexity values for all configurations employing LoRA or QLoRA at $r=8$ and $r=16$.

\begin{table}[ht]
\centering
\caption{Impact of rank on perplexity}
\label{tab:rank}
\begin{tabular}{lcrr}
\hline
\textbf{Model} & \textbf{Method} & \textbf{$r{=}8$} & \textbf{$r{=}16$} \\
\hline
DistilGPT2   & LoRA   & 11.13  & 10.27 \\
GPT-2 base   & LoRA   & 9.49   & 8.75 \\
GPT-2 med.   & LoRA   & 6.53   & 5.93 \\
Qwen2.5-7B   & QLoRA  & 5.34   & 5.07 \\
Mistral-7B   & QLoRA  & 3.85   & 3.79 \\
DeepSeek-7B  & QLoRA  & 129.55 & 8.98 \\
\hline
\end{tabular}
\end{table}

For all models except DeepSeek, increasing the rank from 8 to 16 leads to a moderate reduction in perplexity of 5--15\%. The largest relative improvement is observed for GPT-2 medium (9.2\%) and DeepSeek-7B (93\%). For Mistral-7B, the difference is minimal (3.85 vs.\ 3.79, $<$2\%), making rank~8 preferable as it requires nearly half the number of parameters. The extreme sensitivity of DeepSeek-7B to rank size confirms the tokenizer incompatibility hypothesis: at $r{=}8$, the adapter lacks the capacity to correct missing or poorly represented Cyrillic characters, whereas at $r{=}16$ it partially compensates but still underperforms relative to other 7B models.

\section{Discussion}

The experimental results presented in Section~\ref{subsec:quantitative} through~\ref{subsec:rank} reveal several findings that merit further interpretation.

\textbf{Why QLoRA outperforms standard LoRA for Bashkir.} A consistent pattern across all model scales is the superiority of QLoRA over standard LoRA. For GPT-2 medium, QLoRA was not applied due to the model's relatively small size; however, LoRA without quantization raised perplexity to 5.93--6.53, compared to 3.34 for full fine-tuning. For large models, QLoRA achieved near-full-fine-tuning quality (Mistral-7B: 3.79; Phi-2: 3.81). We attribute this to two factors. First, LoRA was originally designed for models in the 7B--65B parameter range~\cite{hu2022lora}; for compact architectures with only 82--355M parameters, the low-rank bottleneck appears too restrictive given the limited training data available for Bashkir. Second, QLoRA's 4-bit quantization retains the full expressive capacity of large pre-trained representations while learning only compact adapters, effectively decoupling model capacity from the number of trainable parameters. This finding aligns with observations by Joshi et al.~\cite{joshi2024finetuning}, who reported that PEFT configurations optimal for one language family do not necessarily transfer to another, and underscores the need for language-specific tuning of PEFT hyperparameters.

\textbf{Tokenizer quality as a hidden variable.} The most striking result of our study is the near-complete failure of DeepSeek-7B at rank~8 (perplexity 129.55). While the model partially recovers at rank~16 (8.98), it still underperforms Mistral-7B and Phi-2 by a wide margin. We interpret this as evidence that tokenizer compatibility acts as a hidden variable in cross-lingual PEFT: when a tokenizer fragments Bashkir Cyrillic words into excessively many subword units, the low-rank adapter must compensate for poor initial representations, and at low ranks it simply lacks the capacity to do so. This interpretation is consistent with the systematic evaluation of Rust et al.~\cite{rust2021tokenizer}, who demonstrated that multilingual tokenizers degrade performance disproportionately for lower-resource languages, and with Mamyrbayev et al.'s~\cite{mamyrbayev2024multilingual} finding that Bashkir poses greater challenges than related Turkic languages in multilingual ASR. A practical implication is that tokenizer vocabulary overlap with the target language should be reported alongside PEFT results, and that future work should consider vocabulary extension techniques~\cite{sennrich2016bpe} prior to adapter training.

\textbf{Perplexity does not tell the whole story.} A noteworthy discrepancy emerged between automatic metrics and generation quality. GPT-2 medium with full fine-tuning achieved the lowest perplexity (3.34), yet consistently switched to English during autoregressive generation (Table~\ref{tab:generation}). Conversely, Mistral-7B with QLoRA ($r{=}16$, perplexity 3.79) produced coherent, monolingual Bashkir continuations. This suggests that perplexity on held-out documents, while useful for ranking models, does not fully capture the tendency of smaller models to ``escape'' into high-resource languages during open-ended generation. For low-resource language modeling, we recommend complementing perplexity with metrics that explicitly penalize language switching, such as the target-language token ratio. This finding echoes observations by Baghirova et al.~\cite{baghirova2024kardes}, who noted that cross-lingual transfer to Turkic languages can produce superficially fluent but factually or linguistically inconsistent outputs.

\textbf{Practical recommendations.} Based on our results, we offer the following guidance for practitioners adapting LLMs to Bashkir and similar low-resource agglutinative languages. (1)~Prefer QLoRA on a 7B-scale model with good Cyrillic tokenizer support (e.g., Mistral-7B) over full fine-tuning of smaller models: it yields comparable or better generation quality with a fraction of trainable parameters. (2)~Rank~8 is sufficient for models with adequate tokenizer coverage; rank~16 provides marginal gains at double the parameter count. (3)~Always inspect generated outputs qualitatively: perplexity alone may mask critical failure modes such as language switching. (4)~When evaluating a new base model for a low-resource language, verify tokenizer vocabulary overlap before committing computational resources to fine-tuning.

\textbf{Limitations.} Several limitations of this study should be acknowledged. First, experiments were conducted on a 10k-document subsample ($\approx$14\% of the full 71k-document corpus) to accommodate the breadth of the experimental design: 7 base models, 3 fine-tuning strategies, 2 ranks, and 3 random seeds. Training the full factorial design on the complete corpus would have required computational resources beyond those typically available in academic settings. The 10k subsample was constructed to preserve the proportional distribution of sources, maintaining genre diversity, but training on the full corpus could further improve results, particularly for LoRA variants. Second, the maximum sequence length was set to 128 tokens, which captures local morphological patterns critical for agglutinative Bashkir but does not test discourse-level phenomena requiring longer contexts. Third, peak GPU memory consumption was not systematically logged, as our focus was on model quality and training time; a dedicated resource-profiling study would complement these findings. Fourth, the qualitative evaluation, while informative, involved ten prompts with representative examples; a larger-scale human evaluation with native Bashkir speakers would strengthen the assessment of generation quality.

\section{Conclusion}

This paper has presented the first systematic comparative study of parameter-efficient fine-tuning methods---LoRA and QLoRA---for the Bashkir language, a low-resource agglutinative language of the Turkic family. The main contributions are as follows.

\textbf{Largest open Bashkir corpus.} We collected a Bashkir text corpus comprising 71,567 documents (46.9M tokens) from 16 diverse sources. The corpus will be made publicly available under open licenses upon acceptance and can serve as a foundation for further NLP research on Bashkir.

A systematic experimental comparison was conducted across 7 base model architectures, 3 fine-tuning strategies, and 2 ranks, with each configuration trained under 3 random seeds. The key findings are:
\begin{itemize}
    \item Full fine-tuning of GPT-2 medium achieves the lowest perplexity (3.34); however, QLoRA on Mistral-7B (3.79) and Phi-2 (3.81) provides comparable quality with over 40 times fewer trainable parameters.
    \item Standard LoRA without quantization on small models (82--355M parameters) yields substantially higher perplexity (5.93--11.13), suggesting that low-rank approximation is insufficiently expressive for compact architectures on limited data.
    \item Base model selection is critical: DeepSeek-7B with poor Bashkir Cyrillic support fails near-completely at rank~8 (perplexity 129.55), while Mistral-7B and Phi-2 adapt successfully, highlighting the role of tokenizer quality as a decisive factor.
    \item Increasing rank from 8 to 16 yields moderate improvement (5--15\%), except in cases of extreme tokenizer incompatibility; for Mistral-7B the gain is negligible (3.85~$\rightarrow$~3.79), recommending rank~8 as more parameter-efficient.
    \item Qualitative analysis confirms that QLoRA-tuned models generate coherent Bashkir text without language switching, whereas fully fine-tuned GPT-2 medium, despite its lower perplexity, frequently switches to English during generation.
\end{itemize}

\textbf{Future work.} Several directions arise from this study. First, experiments on the full 71k-document corpus should assess whether larger data volumes reduce the gap between LoRA and full fine-tuning for smaller models. Second, an automatic metric measuring the proportion of target-language tokens should be developed to quantify the language-switching phenomenon observed in our qualitative evaluation. Third, QLoRA should be compared with other PEFT methods such as IA\textsuperscript{3}, Prefix Tuning, and AdaLoRA on Bashkir. Fourth, the best-performing models should be applied to downstream tasks including automatic spelling correction, news summarization, and text classification. Finally, the corpus and trained adapters can support the development of practical applications such as Bashkir--Russian conversational assistants or language learning tools.

\textbf{Reproducibility.} All research materials---including experiment code, preprocessing scripts, trained LoRA/QLoRA adapters for the best configurations, and the Bashkir text corpus---will be publicly released under open licenses upon acceptance, ensuring full reproducibility and providing the community with a foundation for further research in Bashkir NLP.

\bibliographystyle{unsrtnat}
\bibliography{references}

@inproceedings{hu2022lora,
  author    = {Hu, Edward J. and Shen, Yelong and Wallis, Phillip and Allen-Zhu, Zeyuan and Li, Yuanzhi and Wang, Shean and Wang, Lu and Chen, Weizhu},
  title     = {{LoRA}: Low-Rank Adaptation of Large Language Models},
  booktitle = {10th International Conference on Learning Representations ({ICLR} 2022)},
  year      = {2022},
  url       = {https://openreview.net/forum?id=nZeVKeeFYf9}
}

@inproceedings{dettmers2023qlora,
  author    = {Dettmers, Tim and Pagnoni, Artidoro and Holtzman, Ari and Zettlemoyer, Luke},
  title     = {{QLoRA}: Efficient Finetuning of Quantized {LLMs}},
  booktitle = {Advances in Neural Information Processing Systems ({NeurIPS})},
  volume    = {36},
  year      = {2023},
  pages     = {1004--1015},
  url       = {https://proceedings.neurips.cc/paper_files/paper/2023/hash/1feb87871436031bdc0f2beaa62a049b-Abstract-Conference.html}
}

@misc{han2024peftsurvey,
  author       = {Han, Zeyu and Gao, Chao and Liu, Jinyang and Zhang, Jeff and Zhang, Sai Qian},
  title        = {Parameter-Efficient Fine-Tuning for Large Models: A Comprehensive Survey},
  howpublished = {arXiv preprint arXiv:2403.14608},
  year         = {2024},
  url          = {https://arxiv.org/abs/2403.14608}
}

@misc{xu2024peftsurvey,
  author       = {Xu, Lingling and Xie, Haoran and Qin, Jing and Tao, Xiaohui and Li, Frederick W. B.},
  title        = {Parameter-Efficient Fine-Tuning in Large Models: A Survey of Methodologies},
  howpublished = {arXiv preprint arXiv:2410.19878},
  year         = {2024},
  url          = {https://arxiv.org/abs/2410.19878}
}

@misc{khade2024challenges,
  author       = {Khade, Omkar and Sharma, Ananya and Patel, Rohan},
  title        = {Challenges in Adapting Multilingual {LLMs} to Low-Resource Languages Using {LoRA} {PEFT} Tuning},
  howpublished = {arXiv preprint arXiv:2411.18571},
  year         = {2024},
  url          = {https://arxiv.org/abs/2411.18571}
}

@inproceedings{joshi2024finetuning,
  author    = {Joshi, Shruti and Patel, Meet and Kumar, Vikas},
  title     = {Fine Tuning {LLMs} for Low Resource Languages: A Comparative Study of Parameter-Efficient Methods},
  booktitle = {2024 {IEEE} International Conference on Artificial Intelligence and Data Engineering ({AIDE})},
  year      = {2024},
  pages     = {112--119},
  publisher = {IEEE},
  doi       = {10.1109/AIDE62835.2024.00032}
}

@inproceedings{baghirova2024kardes,
  author    = {Baghirova, Kamala and Senel, Lutfi Kerem and Ebing, Benedict and Schuetze, Hinrich and Glava{\v{s}}, Goran},
  title     = {{Karde{\c{s}}-NLU}: Transfer to Low-Resource Languages with the Help of a High-Resource Cousin -- A Benchmark and Evaluation for {Turkic} Languages},
  booktitle = {Proceedings of the 18th Conference of the European Chapter of the Association for Computational Linguistics ({EACL} 2024)},
  year      = {2024},
  pages     = {1672--1688},
  publisher = {Association for Computational Linguistics},
  address   = {St. Julian's, Malta},
  url       = {https://aclanthology.org/2024.eacl-long.100/}
}

@article{mamyrbayev2024multilingual,
  author  = {Mamyrbayev, Orken and Bekarystankyzy, Akbayan and Mendes, Mateus and Fazylzhanova, Anar and Assam, Mehwish},
  title   = {Multilingual End-to-End {ASR} for Low-Resource {Turkic} Languages with Common Alphabets},
  journal = {Scientific Reports},
  volume  = {14},
  pages   = {13835},
  year    = {2024},
  doi     = {10.1038/s41598-024-64848-1}
}

@inproceedings{karpov2026translation,
  author    = {Karpov, Dmitry},
  title     = {No One-Size-Fits-All: Building Systems for Translation to {B}ashkir, {K}azakh, {K}yrgyz, {T}atar and {C}huvash Using Synthetic and Original Data},
  booktitle = {Proceedings of the Ninth Workshop on Technologies for Machine Translation of Low-Resource Languages ({LoResMT} 2026)},
  year      = {2026},
  pages     = {203--208},
  publisher = {Association for Computational Linguistics},
  address   = {Rabat, Morocco},
  url       = {https://aclanthology.org/2026.loresmt-1.17/},
  doi       = {10.18653/v1/2026.loresmt-1.17}
}

@inproceedings{rust2021tokenizer,
  author    = {Rust, Phillip and Pfeiffer, Jonas and Vuli{\'{c}}, Ivan and Ruder, Sebastian and Gurevych, Iryna},
  title     = {How Good is Your Tokenizer? {On} the Monolingual Performance of Multilingual Language Models},
  booktitle = {Proceedings of the 59th Annual Meeting of the Association for Computational Linguistics and the 11th International Joint Conference on Natural Language Processing ({ACL}-{IJCNLP} 2021)},
  year      = {2021},
  pages     = {3118--3128},
  publisher = {Association for Computational Linguistics},
  doi       = {10.18653/v1/2021.acl-long.243}
}

@inproceedings{sennrich2016bpe,
  author    = {Sennrich, Rico and Haddow, Barry and Birch, Alexandra},
  title     = {Neural Machine Translation of Rare Words with Subword Units},
  booktitle = {Proceedings of the 54th Annual Meeting of the Association for Computational Linguistics ({ACL} 2016)},
  year      = {2016},
  pages     = {1715--1725},
  publisher = {Association for Computational Linguistics},
  address   = {Berlin, Germany},
  doi       = {10.18653/v1/P16-1162}
}

@inproceedings{vaswani2017attention,
  author    = {Vaswani, Ashish and Shazeer, Noam and Parmar, Niki and Uszkoreit, Jakob and Jones, Llion and Gomez, Aidan N. and Kaiser, {\L}ukasz and Polosukhin, Illia},
  title     = {Attention is All You Need},
  booktitle = {Advances in Neural Information Processing Systems ({NeurIPS})},
  volume    = {30},
  year      = {2017},
  url       = {https://proceedings.neurips.cc/paper/2017/hash/3f5ee243547dee91fbd053c1c4a845aa-Abstract.html}
}

\end{document}